\documentclass[conference]{IEEEtran}

\IEEEoverridecommandlockouts

\usepackage{graphicx}
\usepackage{amsmath,amssymb} 
\usepackage{lineno}
\usepackage{color}

\usepackage{multirow}
\usepackage{hyperref}
\usepackage{flushend}

\def\b{\ensuremath\boldsymbol}

\usepackage[algo2e,linesnumbered,lined,boxed,vlined]{algorithm2e}

\ifCLASSINFOpdf

\else

\fi

\usepackage{eso-pic} 

\begin{document}

\AddToShipoutPictureBG*{%
  \AtPageUpperLeft{%
    \setlength\unitlength{1in}%
    \hspace*{\dimexpr0.5\paperwidth\relax}
    \makebox(0,-0.75)[c]{\normalsize Accepted for presentation at the 25th International Conference on Pattern Recognition (ICPR), IEEE, 2020.}
    }}

\title{Batch-Incremental Triplet Sampling for Training Triplet Networks Using  Bayesian Updating Theorem}

\author{\IEEEauthorblockN{Milad Sikaroudi\IEEEauthorrefmark{1}$\ddagger$, Benyamin Ghojogh\IEEEauthorrefmark{2}$\ddagger$, Fakhri Karray\IEEEauthorrefmark{2},~\IEEEmembership{Fellow,~IEEE}, \\
Mark Crowley\IEEEauthorrefmark{2},~\IEEEmembership{Member,~IEEE}, H.R. Tizhoosh\IEEEauthorrefmark{1},~\IEEEmembership{Senior Member,~IEEE} \thanks{$\ddagger$ The first two authors contributed equally to this work.}
}
\IEEEauthorblockA{
\IEEEauthorrefmark{1}Kimia Lab, University of Waterloo, Canada \\
\IEEEauthorrefmark{2}Department of Electrical and Computer Engineering, University of Waterloo, Canada \\
Emails: \{msikaroudi, bghojogh, karray, mcrowley, tizhoosh\}@uwaterloo.ca
}}

\maketitle

\begin{abstract}
Variants of Triplet networks are robust entities for learning a discriminative embedding subspace. There exist different triplet mining approaches for selecting the most suitable training triplets. Some of these mining methods rely on the extreme distances between instances, and some others make use of sampling. However, sampling from stochastic distributions of data rather than sampling merely from the existing embedding instances can provide more discriminative information. In this work, we sample triplets from distributions of data rather than from existing instances. We consider a multivariate normal distribution for the embedding of each class. Using Bayesian updating and conjugate priors, we update the distributions of classes dynamically by receiving the new mini-batches of training data. The proposed triplet mining with Bayesian updating can be used with any triplet-based loss function, e.g., \textit{triplet-loss} or Neighborhood Component Analysis  (NCA) loss. Accordingly, Our triplet mining approaches are called Bayesian Updating Triplet (BUT) and Bayesian Updating NCA (BUNCA), depending on which loss function is being used. Experimental results on two public datasets, namely MNIST and histopathology colorectal cancer (CRC), substantiate the effectiveness of the proposed triplet mining method. 
\end{abstract}

\IEEEpeerreviewmaketitle

\section{Introduction}\label{section_introduction}

Variants of Siamese networks contain several, typically two \cite{hadsell2006dimensionality} or three \cite{schroff2015facenet,hoffer2015deep}, sub-networks sharing their weights. The Siamese topologies are robust networks for learning a discriminative embedding space, i.e., explicit metric space, between the classes of data \cite{chang2016image}. One of these variants is the triplet network in which anchor, positive and negative triplets are used for decreasing and increasing the distance of anchor-positive and anchor-negative pairs, respectively \cite{schroff2015facenet}, resulting in increasing and decreasing the inter- and intra-class variances of data \cite{ghojogh2020fisher}. 
Two popular forms of loss function for training triplets are \textit{triplet-loss} \cite{schroff2015facenet} and the softmax form \cite{ye2019unsupervised}. Some examples for the latter are Neighborhood Component Analysis (NCA) \cite{goldberger2005neighbourhood} and proxy-NCA \cite{movshovitz2017no}. 

Apart from the loss functions, there is another degree of freedom, which is how the triplets are sampled. It is shown in \cite{wu2017sampling} that sampling of the triplets also matters in learning deep embeddings. Hence, proposing a decent sampling strategy has not less importance than a novel loss function. 
In other words, with triplet networks, drawing more informative and stable triplets from the pool of samples will lead to qualitatively more salient embeddings. 

There are already some triplet mining strategies in the literature.
Instead of using all the triplets in a mini-batch of data, i.e., Batch All (BA) \cite{ding2015deep}, one can mine the triplets as in Batch Semi-Hard (BSH) \cite{schroff2015facenet} and Batch Hard (BH) \cite{hermans2017defense}.
Some mining methods, such as Easy Positive (EP) \cite{xuan2020improved}, concentrate on the extreme distances of samples. However, some other triplet mining methods use the concept of sampling from the available triplets in a mini-batch of the data \cite{wu2017sampling}. 

In this work, we aim to draw the positive and negative samples for every anchor instance in a dynamic manner. The main idea is to sample the positive and negative instances of triplets for every anchor in a mini-batch of data from some distributions rather than from the embedded data points themselves. This gives the triplet network more opportunity to explore the embedding space for increasing and decreasing the inter- and intra-class variances because the triplet information is not restricted to only the embedded data but is instead  stochastic. That is while the related work on triplet sampling samples the triplets from the existing embedded data instances \cite{wu2017sampling}, it does not use the stochastic information of the embedding space. We assume a multivariate normal distribution for the embedded data instances of every class. These distributions are updated dynamically by receiving new streaming embedded data for the different classes. For this dynamic updating, we leverage the theory of Bayesian distribution updating \cite{jaffray1992bayesian,gelman2013bayesian} and conjugate priors \cite{murphy2007conjugate,jordan2010conjugate}. 
Sampling from dynamic distributions makes the task of sampling not only more robust to outliers but also more amenable to available data. 
The proposed approaches are called Bayesian Updating for \textit{triplet-loss} (BUT) and Bayesian Updating for NCA loss (BUNCA). 

The rest of the paper is organized as follows: Section \ref{section_background} introduces the necessary background on Bayesian updating and conjugate priors. The dynamic triplet sampling for training triplet networks is proposed in Section \ref{section_dynamic_sampling}. We report and discuss the experimental results in Section \ref{section_experiments}. Finally, Section \ref{section_conclusion} concludes the paper and highlights the possible future work. 

\section{Background on Bayesian Updating}\label{section_background}
In this section, we describe the Bayesian updating and the conjugate priors. As well, we briefly review relevant distributions to lay the foundation for dynamic triplet sampling of our approach. 

\subsection{Bayesian Updating}
Let $X$ and $\theta$ be two random variables where $\theta$ is a parameter of the distribution of $X$. According to Bayes' rule, we have
\begin{align}\label{equation_Bayes_rule}
\mathbb{P}(\theta | X) = \frac{\mathbb{P}(X | \theta)\, \mathbb{P}(\theta)}{\mathbb{P}(X)} \implies \mathbb{P}(\theta | X) \propto \mathbb{P}(X | \theta)\, \mathbb{P}(\theta),
\end{align}
which shows the relation of the posterior $\mathbb{P}(\theta | X)$, likelihood $\mathbb{P}(X | \theta)$, and prior $\mathbb{P}(\theta)$. 
Given some data $X$ and the prior over the parameter of interest $\theta$, we want to find the posterior using Eq. (\ref{equation_Bayes_rule}). This is the basic idea behind \textit{Bayesian updating} in which the posterior over the parameter of interest is updated after receiving some new data, i.e., using the new data $X$, we have $\mathbb{P}(\theta) \mapsto \mathbb{P}(\theta | X)$ \cite{jaffray1992bayesian}. 

\subsection{Conjugate Priors}

If the posterior distribution $\mathbb{P}(\theta | X)$ and the prior distribution $\mathbb{P}(\theta)$ are in the same probability distribution family, they are called \textit{conjugate distributions} and the prior is the \textit{conjugate prior} for the likelihood $\mathbb{P}(X | \theta)$ \cite{gelman2013bayesian}. 

Assume there already exist some data, denoted by $X^0$, and some new data, $X'$, are received. 
The existing data $X^0$ has a distribution with some parameter(s) $\theta$. 
The posterior of the parameter of interest, i.e., $\mathbb{P}(\theta | X)$, can be updated using the new data. Hence, this can be used to update the parameter(s) of the distribution of $X$ using the newly received data \cite{jordan2010conjugate}. 

Let the data $X$ have a multivariate normal (or Gaussian) distribution, so its likelihood is $\mathbb{P}(X | \theta)$. Assume both the mean and covariance of likelihood are considered as random variables, so $\theta$ includes mean and covariance. Using the new data $X'$, we want to update the parameters, mean and covariance, of the normal distribution. In this case, the likelihood $\mathbb{P}(X | \theta)$ has a multivariate normal distribution, and for updating the posterior, we should use the conjugate prior for the likelihood. The conjugate prior distribution for the multivariate normal distribution with both random mean and covariance is the normal-inverse-Wishart distribution \cite{murphy2007conjugate}. In our analysis, we also require the skewed generalized Student-$t$ distribution. 

\subsection{Relevant Distributions}

\textbf{Multivariate Normal Distribution}: The Probability Density Function (PDF) of the \textit{multivariate normal distribution} is defined as \cite{gelman2013bayesian}
\begin{align}
&X \sim \mathcal{N}(\b{\mu}, \b{\Sigma}) := \nonumber \\
& \frac{1}{\sqrt{(2 \pi)^{d}\, |\b{\Sigma}|}} \exp\Big(\!\!-\frac{1}{2} (\b{x} - \b{\mu})^\top \b{\Sigma^{-1} (\b{x} - \b{\mu})}\Big),
\end{align}
where $d$ is the dimensionality of data, $|\cdot|$ denotes the determinant of matrix, and $\b{x} \in \mathbb{R}^d$, $\b{\mu} \in \mathbb{R}^d$, and $\b{\Sigma} \in \mathbb{R}^{d \times d}$ are the data, mean, and covariance of data, respectively. The mean and covariance of the normal distribution can be estimated by the sample mean and sample covariance matrix, respectively. 

\textbf{Wishart and Inverse Wishart Distributions}: The PDF of the \textit{Wishart distribution} is defined as \cite{gelman2013bayesian}
\begin{align}
&X \sim \mathcal{W}_d(\b{V}, \nu) := \nonumber \\
& \frac{1}{2^{(\nu d)/2}\, |\b{V}|^{\nu/2}\, \Gamma_d(\frac{\nu}{2})}\, |\b{x}|^{(\nu - d - 1)/2}\, \exp(-\frac{1}{2}\textbf{tr}(\b{V}^{-1} \b{x})),
\end{align}
where $\nu$ is the degrees of freedom (which should be $\nu \geq d$), $\mathbb{R}^{d \times d} \ni \b{V} \succ 0$ is the scale matrix, $\textbf{tr}(\cdot)$ denotes the trace of matrix, and $\Gamma_d(\cdot)$ is the \textit{multivariate gamma function} \cite{gupta2018matrix}:
\begin{align}
\Gamma_d(a) := \int_{\b{S} \succ 0} \exp\big(\!-\textbf{tr}(\b{S})\big)\, |\b{S}|^{a - (d+1)/2}\, d\b{S}.
\end{align}
Consider a variable with Wishart distribution, i.e., $Z \sim \mathcal{W}_d(\b{V}, \nu)$. Then, the variable $X = Z^{-1}$ has the \textit{inverse Wishart distribution} whose PDF is defined as \cite{gelman2013bayesian}:
\begin{align}
&X \sim \,\mathcal{W}_d^{-1}(\b{\Psi}, \nu):= \nonumber \\ 
&  \frac{|\b{\Psi}|^{\nu/2}}{2^{(\nu d)/2}\, \Gamma_d(\frac{\nu}{2})}\, |\b{x}|^{-(\nu + d + 1)/2}\, \exp(-\frac{1}{2}\textbf{tr}(\b{\Psi} \b{x}^{-1})),
\end{align}
where $\mathbb{R}^{d \times d} \ni \b{\Psi} \succ 0$ is the scale matrix and we have $\b{\Psi} = \b{V}^{-1}$ \cite{marola1979multivariate}. From the moments of the inverse Wishart distribution, the mean of a random variable $X \sim \mathcal{W}_d^{-1}(\b{\Psi}, \nu)$ is defined as follows \cite{von1988moments}:
\begin{align}\label{equation_expectation_inverse_Wishart}
\mathbb{E}(X) = \frac{\b{\Psi}}{\nu - d - 1}, \quad \forall\, \nu > d + 1.
\end{align}

\textbf{Skewed Generalized Student-$t$ Distribution}: 
The PDF of the \textit{Student-$t$ distribution} is defined as \cite{gelman2013bayesian}
\begin{align}
X \sim t_\nu := \frac{\Gamma(\frac{\nu + 1}{2})}{\sqrt{\nu \pi}\, \Gamma(\frac{\nu}{2})}\, (1 + \frac{x^2}{\nu})^{-(\nu + 1) / 2}, 
\end{align}
where $\nu > 0$ is a degree of freedom and $\Gamma(\nu) := (\nu - 1)!$ is the Gamma function. The Student-$t$ distribution can be generalized which is called the \textit{skewed generalized Student-$t$ distribution} whose PDF is defined as \cite{murphy2007conjugate,theodossiou1998financial}
\begin{align}
X\!\sim\!t_\nu(\mu, \sigma^2)\!:=\! \frac{\Gamma(\frac{\nu + 1}{2})}{\sqrt{\nu \pi}\, \sigma\, \Gamma(\frac{\nu}{2})}\, \Big(1\!+\!\frac{1}{\nu} \big(\frac{x-\mu}{\sigma}\big)^2 \Big)^{-\frac{(\nu + 1)}{2}}, 
\end{align}
where $\mu$ and $\sigma^2$ are the mean and variance, respectively. 
The generalized Student-$t$ distribution can be $d$-dimensional multivariate {\cite[Definition 2]{papastathopoulos2013generalised}}: 
\begin{align}
&X \sim t_\nu(\b{\mu}, \b{\Sigma}):= \nonumber \\
& \frac{\Gamma(\frac{\nu + d}{2})}{(\nu \pi)^{\nu/2}\, \Gamma(\frac{\nu}{2})}\, \Big(1 + \frac{1}{\nu} (\b{x} - \b{\mu})^\top \b{\Sigma}^{-1} (\b{x} - \b{\mu})\Big)^{-\frac{(\nu + 1)}{2}}, 
\end{align}
where $\b{\mu} \in \mathbb{R}^d$ and $\mathbb{R}^{d \times d}$ are the mean and covariance, respectively. 
The mean of the skewed generalized Student-$t$ distribution is $\mathbb{E}(X) = \b{\mu}$ \cite{murphy2007conjugate}. 

\textbf{Normal-Inverse-Wishart Distribution:} As was mentioned before, the prior distribution for the multivariate normal distribution with both mean and covariance as random variables is the inverse Wishart distribution. 
Recall that we have some existing data denoted by $X^0$. We show the set of existing data vectors by $\{\b{x}^0_i\}_{i=1}^{n_0}$ where $n_o$ is the sample size of the existing data. Assume that data have a multivariate normal distribution $X \sim \mathcal{N}(\b{\mu}, \b{\Sigma})$. 
Let $\mathbb{R}^d \ni \b{\mu}^0 := (1/n_0) \sum_{i=1}^{n_0} \b{x}_i^0$ and $\mathbb{R}^d \ni \b{\mu}' := (1/n') \sum_{i=1}^{n'} \b{x}'_i$ denote the sample mean of the existing and new data, respectively. Likewise, $\mathbb{R}^{d \times d} \ni \b{\Sigma}^0 := (1/n_0) \sum_{i=1}^{n_0} (\b{x}^0_i - \b{\mu}^0) (\b{x}^0_i - \b{\mu}^0)^\top$ and $\mathbb{R}^{d \times d} \ni \b{\Sigma}' := (1/n') \sum_{i=1}^{n'} (\b{x}'_i - \b{\mu}') (\b{x}'_i - \b{\mu}')^\top$ are the sample covariance matrix over the existing and new data, respectively. 

The prior of covariance is $\b{\Sigma} \sim \mathcal{W}^{-1}_d(\b{\Sigma}'^{-1}, n')$ and the distribution of mean given covariance is $\b{\mu} | \b{\Sigma} \sim \mathcal{N}(\b{\mu}', (1/n') \b{\Sigma})$ \cite{gelman2013bayesian,murphy2007conjugate}. 
The joint distribution of the mean and covariance is the \textit{Normal-Inverse-Wishart (NIW) distribution} \cite{gelman2013bayesian,murphy2007conjugate}: 
\begin{equation}
\begin{aligned}
&\mathbb{P}(\b{\mu}, \b{\Sigma}) = \text{NIW}(\b{\mu}', \nu'_1, \b{\Sigma}', \nu'_2) := \\
&~  \frac{|\b{\Sigma}'|^{\nu'_2/2} |\b{\Sigma}|^{-((\nu'_2+d)/2 + 1)}}{2^{(\nu'_2 d)/2} \Gamma_d(\frac{\nu'_2}{2}) (\frac{2 \pi}{\nu'_1})^{d/2}} \times \\
&~~~ \exp\Big(\!\! -\!\frac{1}{2}\textbf{tr}(\b{\Sigma'} \b{\Sigma}^{-1})\!-\! \frac{\nu'_1}{2} (\b{\mu}\!-\! \b{\mu}')^\top \b{\Sigma}^{-1} (\b{\mu}\!-\!\b{\mu}') \Big),
\end{aligned}
\end{equation}
where $\nu'_1$ and $\nu'_2$ are the sample sizes of new data used for calculating the new mean and covariance matrix. In this work, we have $\nu'_1 = \nu'_2 = n'$. 

The posterior of mean and covariance of data is again a NIW distribution \cite{gelman2013bayesian,murphy2007conjugate}:
\begin{align}
& \mathbb{P}(\b{\mu}, \b{\Sigma}\, |\, \b{x}^0, \b{\mu}', \nu'_1, \b{\Sigma}', \nu'_2) \nonumber \\
&~~~~~~~~~~ = \text{NIW}\big(\b{\mu}, \b{\Sigma}\, \big|\, \b{\eta}, \nu'_1 + n_0, \b{\Upsilon}, \nu'_2 + n_0 \big), \\
& \mathbb{R}^d \ni \b{\eta} := \frac{\nu'_1 \b{\mu}' + n_0 \b{\mu}^0}{\nu'_1 + n_0}, \label{equation_eta} \\
& \mathbb{R}^{d \times d} \ni \b{\Upsilon}\! := \! \nu'_2 \b{\Sigma}' + n_0 \b{\Sigma}^0 + \frac{\nu'_1 n_0}{\nu'_1 + n_0} (\b{\mu}^0 - \b{\mu}') (\b{\mu}^0 - \b{\mu}')^\top. \label{equation_Upsilon}
\end{align}
The marginal distributions of mean and covariance of data are \cite{gelman2013bayesian,murphy2007conjugate}:
\begin{align}
&\mathbb{P}(\b{\mu}\, |\, \b{x}^0) = t_{\nu'_2+n_0-d+1}\Big(\b{\eta}, \frac{\b{\Upsilon}}{(\nu'_1 + n_0)(\nu'_2+n_0-d+1)}\Big),  \label{equation_marginal_mean} \\
&\mathbb{P}(\b{\Sigma}\, |\, \b{x}^0) = \mathcal{W}_d^{-1}(\b{\Upsilon}^{-1}, \nu'_2 + n_0), \label{equation_marginal_covariance} 
\end{align}
respectively. The Eqs. (\ref{equation_marginal_mean}) and (\ref{equation_marginal_covariance}) can be used to update the parameters of a multivariate normal distribution upon receiving the new data. 


\section{Dynamic Triplet Sampling for Training Triplet Networks}\label{section_dynamic_sampling}

\subsection{Preliminaries and Notations}

Consider a $q$-dimensional training dataset $\{\b{z}_i\}_{i=1}^n$ where $\b{z}_i \in \mathbb{R}^q$. The class labels of instances are $\{y_i\}_{i=1}^n$. 
Suppose we have $c$ number of classes in the dataset. 
We use the mini-batch (of size $b$) stochastic gradient descent for training the network. 
Let $n^j$ denote the training sample size per class in a mini-batch. We show the $i$-th training instance of the $j$-th class in a mini-batch by $\b{z}'^j_i$. 
Let $\b{x}'^j_i \in \mathbb{R}^d$ denote the embedding of $\b{z}'^j_i$ by the triplet network where the dimensionality of embedding space is $d$. 

The data for each class are accumulated by receiving new mini-batches of data. 
Let $n_0^j$ denote the sample size of accumulated data for the $j$-th class so far. 
The sample size per $j$-th class in a mini-batch is denoted by $n'^j$. In this work, we have $n'^1 = \dots = n'^c = n' = \lceil b / c \rceil$ and $n_0^1 = \dots = n_0^c = n_0$ because we take the same sample size per class in the mini-batch. This $n'$ is the sample size of new incoming data per class in every mini-batch. 
The accumulated data for the $j$-th class so far are denoted by $\b{x}^{0,j}$. 
Also, $\b{\mu}^j$ and $\b{\Sigma}^j$ are the mean and covariance of the distribution of the $j$-th class, respectively. 

\subsection{Sampling Algorithm}

We assume a multivariate normal distribution for the embedded data of every class. This assumption makes sense according to the central limit theorem \cite{hazewinkel2001central} and the fact that the normal distribution is the most common continuous distribution. 
In the first batch, where there is not already any embedding of training data, we use Maximum Likelihood Estimation (MLE) to  estimate the distribution parameters. The mean and covariance of the embedded data of every class are estimated by the sample mean and covariance matrix, respectively. 

In later batches after the first batch, we do have some existing data per class, denoted by $n_0^j, \forall j$. 
According to Bayesian updating, the mean and covariance of distribution of every class are updated by Eqs. (\ref{equation_marginal_mean}) and (\ref{equation_marginal_covariance}), respectively. We update the mean and covariance matrix of the distribution of every class by the expectation of Eqs. (\ref{equation_marginal_mean}) and (\ref{equation_marginal_covariance}) which are the generalized Student-$t$ and the inverse Wishart distributions, respectively.
According to the expectations of these two distributions which were introduced in Section \ref{section_background}, the updates of mean and covariance of the $j$-th class can be given as 
\begin{align}
& \b{\mu}^{0,j} \leftarrow \mathbb{E}(\b{\mu}^j\, |\, \b{x}^{0,j}) = \b{\eta}^j \overset{(\ref{equation_eta})}{=} \frac{n' \b{\mu}'^j + n_0 \b{\mu}^{0,j}}{n' + n_0}, \\
& \b{\Sigma}^{0,j} \leftarrow \mathbb{E}(\b{\Sigma}^j\, |\, \b{x}^{0,j}) \overset{(\ref{equation_expectation_inverse_Wishart})}{\!=\!} \frac{\b{\Upsilon}^{-1}}{n'\!+\! n_0\! -\! d \!-\! 1},\! \forall\, n'\! +\! n_0\! >\! d\! +\! 1,
\end{align}
where, in Eq. (\ref{equation_Upsilon}), we use $\nu'_1 = \nu'_2 = n'$ and calculate $\b{\mu}'^j$, $\b{\mu}^{0,j}$, $\b{\Sigma}'^j$, and $\b{\Sigma}^{0,j}$ by sample mean and sample covariance matrix using the new batch of data. 
Note that for $n' + n_0 \leq d + 1$ which is in very first mini-batches of first epoch, we update the covariance matrix by MLE. 

The proposed dynamic triplet sampling is summarized in Algorithm \ref{algorithm_dynamic_sampling}. 
The mean and covariance of every class are estimated by MLE at the initial batch. In the following batches, Bayesian updating is exploited for updating the mean and covariance of classes. 
After the means and covariances are updated, we sample the triplets. For every instance of a batch, considered as an ``anchor'', a negative instance is sampled from each different class resulting in $(c-1)$ negatives per anchor. Accordingly, $(c-1)$ positive instances are also sampled from the same class of anchor. Overall, $(b \times (c-1))$ triplets are sampled in every mini-batch while the distributions of classes are being updated dynamically. 

\SetAlCapSkip{0.5em}
\IncMargin{0.8em}
\begin{algorithm2e}[!t]
\DontPrintSemicolon
    \textbf{Procedure: } TrainTripletNetwork($\{\b{z}_i\}_{i=1}^n$, $\{y_i\}_{i=1}^n$)\;
    \textbf{Input: } training data: $\{\b{z}_i\}_{i=1}^n$, training labels: $\{y_i\}_{i=1}^n$\;
    \For{all required epochs}{
        \For{all batches in epoch}{
            $\{\b{x}_i\}_{i=1}^b \leftarrow$ Feed $\{\b{z}_i\}_{i=1}^b$ to the triplet network\;
            \For{class $j$ from $1$ to $c$}{
                \uIf{it is first mini-batch}{
                    $\b{\mu}^{0,j} := (1 / n') \sum_{i=1}^{n'} \b{x}'^j_i$\;
                    $\b{\Sigma}^{0,j} := (1 / n') \sum_{i=1}^{n'} (\b{x}'^j_i - \b{\mu}^{0,j}) (\b{x}'^j_i - \b{\mu}^{0,j})^\top$\;
                }
                \Else{
                    $\b{\mu}'^{j} := (1 / n') \sum_{i=1}^{n'} \b{x}'^j_i$\;
                    $\b{\mu}^{0,j} := (n' \b{\mu}'^j + n_0 \b{\mu}^{0,j})/(n' + n_0)$\;
                    \uIf{$n' + n_0 > d + 1$}{
                        $\b{\Upsilon} := n' \b{\Sigma}'^j + n_0 \b{\Sigma}^{0,j} + \frac{n' n_0}{n' + n_0} (\b{\mu}^{0,j} - \b{\mu}'^j) (\b{\mu}^{0,j} - \b{\mu}'^j)^\top$\;
                        $\b{\Sigma}^{0,j} := \b{\Upsilon}^{-1} / (n' + n_0 - d - 1)$\;
                    }
                    \Else{
                        $\b{\Sigma}^{0,j} := (1 / n') \sum_{i=1}^{n'} (\b{x}'^j_i - \b{\mu}'^{j}) (\b{x}'^j_i - \b{\mu}'^{j})^\top$\;
                    }
                }
            }
            \For{instance $i$ from $1$ to $b$}{
                anchor $\leftarrow \b{x}_i$\;
                \For{class $j$ from $1$ to $c$}{
                    \uIf{$j = y_i$}{
                        Sample $(c-1)$ positive instances $\sim \mathcal{N}(\b{\mu}^{0,j}, \b{\Sigma}^{0,j})$\;
                    }
                    \Else{
                        Sample a negative instance $\sim \mathcal{N}(\b{\mu}^{0,j}, \b{\Sigma}^{0,j})$\;
                    }
                }
            }
            Minimize the \textit{triplet}/NCA loss with the $(b \times (c-1))$ triplets.\;
        }
    }
\caption{Dynamic Triplet Sampling with Bayesian Updating}\label{algorithm_dynamic_sampling}
\end{algorithm2e}
\DecMargin{0.8em}

\subsection{Optimization of the Loss Functions}

In a mini-batch, let the anchor, positive, and negative instances be indexed by $i$, $k$, $\ell$, respectively. Using $b \times (c-1)$ sampled triplets, the \textit{triplet-loss} function can be employed to train the triplet network \cite{schroff2015facenet}:
\begin{align}
\text{minimize} \sum_{i=1}^{b} \sum_{k=1}^{c-1} \sum_{\ell=1}^{c-1}\! \Big[m\!+\!\|\b{x}_{i}\!- \!\b{x}_{k}\|_2^2 \!-\! \|\b{x}_{i}\!-\!\b{x}_{\ell}\|_2^2 \Big]_+,
\end{align}
where $[\cdot]_+ := \max(\cdot,0)$ denotes the standard Hinge loss and $m$ is a small margin (e.g., $0.25$).
When dynamic triplet sampling is used with the triplet loss, we call this Bayesian Updating for \textit{triplet-loss} (BUT).

As was mentioned before, the \textit{triplet-loss} should increase and decrease the inter- and intra-class variances to have a discriminating embedding space for classes of data. This intuition can also be implemented in a softmax form \cite{ye2019unsupervised} which is referred to as NCA \cite{goldberger2005neighbourhood}. We can use this form to train the network:
\begin{align}
\text{minimize } -\sum_{i=1}^{b} \sum_{k=1}^{c-1} \ln\! \Big(\frac{\exp(-\|\b{x}_i - \b{x}_k\|_2^2)}{\sum_{\ell=1}^{c-1} \exp(-\|\b{x}_i - \b{x}_\ell\|_2^2)}\Big).
\end{align}
We name using dynamic triplet sampling with the NCA loss function Bayesian Updating for NCA loss (BUNCA). 

\section{Experiments}\label{section_experiments}

\begin{figure*}[!t]
\centering
\includegraphics[width=\textwidth]{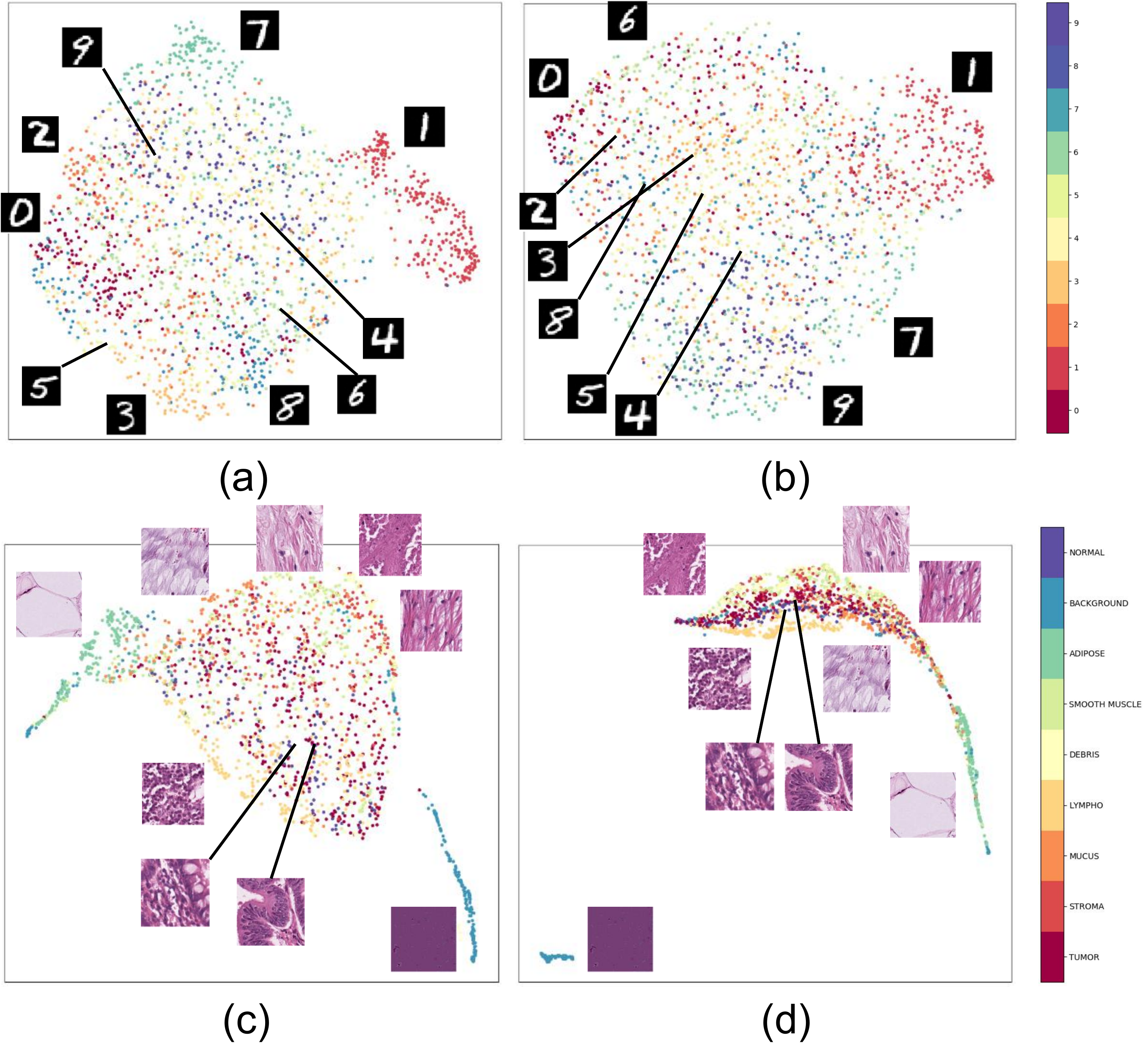}
\caption{2D visualization of test embeddings: (a) MNIST using BUT, (b) MNIST  using BUNCA, (c) CRC using BUT, and (d) CRC using BUNCA.}
\label{figure_embeddings}
\end{figure*}

\begin{figure*}[!t]
\centering
\includegraphics[width=7in]{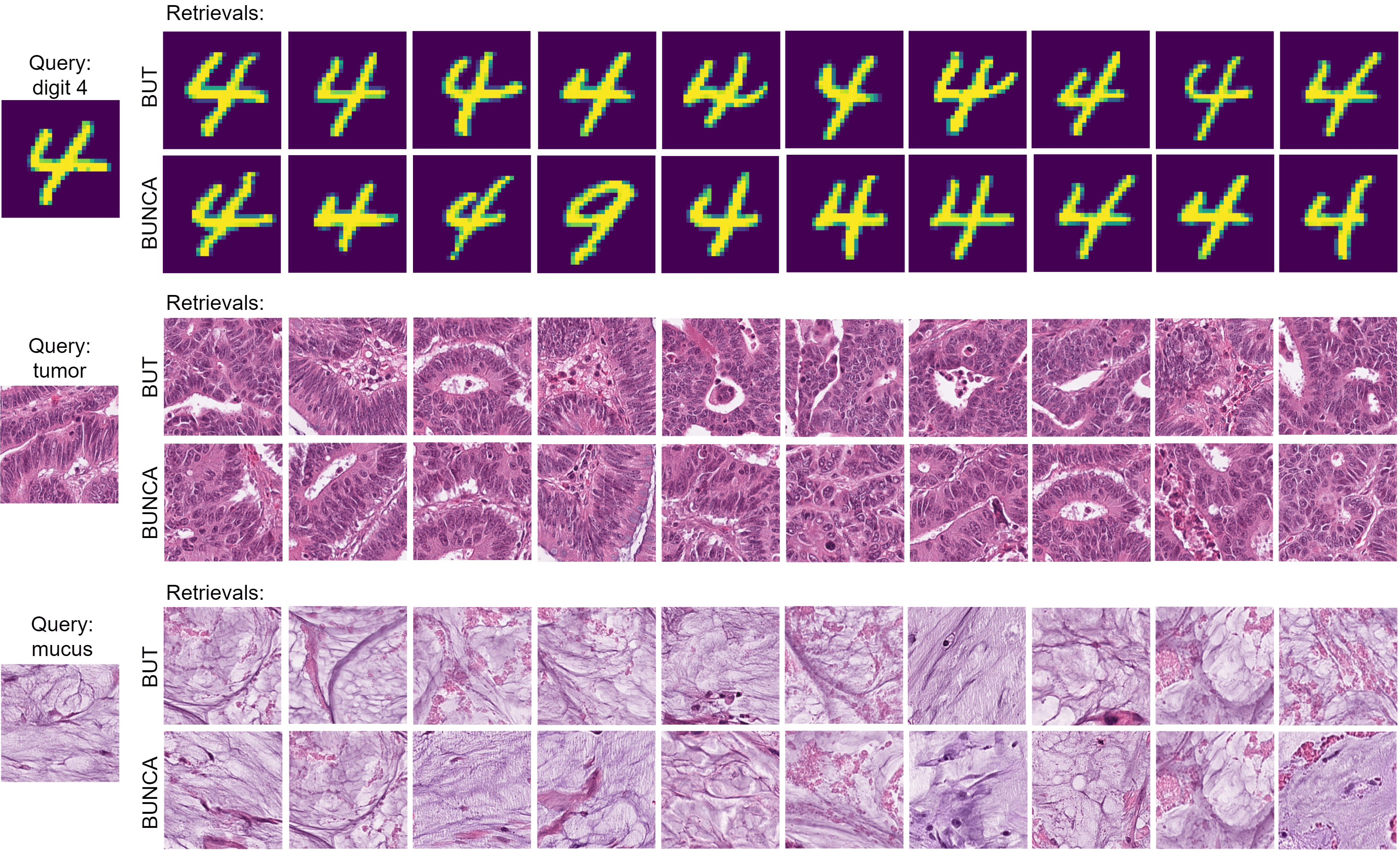}
\caption{Image retrieval in the embedded spaces learned using the BUT and BUNCA approaches. The retrievals are sorted from left to right.}
\label{figure_query}
\end{figure*}

\subsection{Datasets}

We used two different datasets in our  experiments. The first dataset is the MNIST digits data \cite{lecun1998gradient} with 60,000 training instances and 10,000 test instances of size $28 \times 28$ pixels.  
The second dataset we used is the large colorectal cancer (CRC) histopathology dataset \cite{katherDataset,kather2019predicting} with 100,000 stain-normalized image patches of size $224\! \times\! 224$ pixels. The large CRC dataset includes nine classes of tissues, namely adipose, background, debris, lymphocytes, mucus, smooth muscle, normal colon mucosa (normal), cancer-associated stroma, and colorectal adenocarcinoma epithelium (tumor). Note that literature has shown the effectiveness of triplet variants networks for histopathology data, both with \textit{triplet-loss} \cite{sikaroudi2020supervision} and with NCA loss \cite{teh2020learning}; this shows the importance of validating our approaches on this domain.

\subsection{Experimental Setup}

For the MNIST dataset, we split the training data into $70\%$ and $30\%$ portions for training and validation sets. The test set with 10,000 images was used for the test. 
The CRC data were split into training, validation, and test sets with $70\%$, $15\%$, and $15\%$ portions, respectively. 
We used ResNet-18 network \cite{he2016deep} as the backbone of \textit{triplet} network. Using the validation set, early stopping \cite{caruana2001overfitting} was employed, and the maximum number of epochs was set to $50$. 
The batch size was $50$ and $45$ for the MNIST and CRC data, respectively, where every batch contains five instances per class (i.e., $n'=5$).
The learning rate was set to $10^{-5}$, and the dimensionality of the embedding space was $128$. 

\subsection{Visualization of Embedding Spaces}

The 2D visualization of spaces was performed using the Uniform Manifold Approximation and Projection (UMAP) \cite{mcinnes2018umap} applied to the embedded data. Figure \ref{figure_embeddings} illustrates the embedding of test sets of the MNIST and CRC data using the BUT and BUNCA sampling methods. As apparent in this figure, the learned embedding spaces are interpretable. In embeddings of MNIST data, the similar digits, in the style of writing, fall close to one another. Closely embedded digits by BUT (see Fig. \ref{figure_embeddings}-a) are the digits 1 and 7, 7 and 9, 3 and 8, and 4 (second style of writing) and 9. Likewise, closely embedded digits by BUNCA (see Fig. \ref{figure_embeddings}-b) are the digits 0 and 6, 1 and 7, 7 and 9, 3 and 8, and 2 and 3 (because continuing the underneath curve of 2 results in 3). 

The embedding spaces for the histopathology data are also meaningful. The histopathology patches with similar patterns have been embedded close to each other as expected. In embedding using the BUT approach (see Fig. \ref{figure_embeddings}-c), the patches are embedded from smoothest to roughest patterns in a circular manner. These patches, with smoothest to roughest \cite{kather2016multi} patterns, are adipose (with thin stripes of fat), mucus, smooth muscle, debris, stroma, tumor, normal, and lymphocyte (with a rough pattern). Moreover, the background patch with no pattern is separated from the tissues, as expected. 
In embedding using the BUNCA approach (see Fig. \ref{figure_embeddings}-d), the patches with a considerable amount of roughness are embedded closely. For example, adipose, mucus, stroma, and smooth muscle, which are smoother, fall close to each other while tumor, normal, lymphocyte, and debris, with diverse patterns, are embedded close to each other. Again, the background patches are embedded far from the tissue types. 
The meaningfulness of the learned embedded spaces shows the effectiveness of the proposed BUT and BUNCA approaches. 

\subsection{Query Retrieval}

For the evaluation of the embedding space, one can see the embedded instances as a database where nearby cases can be retrieved as matched cases for a query instance. The retrievals are extracted using the nearest neighbors in the embedding space. Because of representation learning, the retrievals are expected to be similar to the query in terms of pattern. In Fig. \ref{figure_query}, we illustrate the top ten retrievals for query examples for both MNIST and histopathology data. 
The retrievals in the embedding spaces using both BUT and BUNCA approaches are shown to visually verify the similarity matching. 

\subsubsection{Retrieval of Digit Images}

In Fig. \ref{figure_query}, the retrievals for a digit 4 with the second style of writing are depicted. As expected, the retrievals are very similar to the pattern of the query image. Compared to the last retrievals, the first retrievals are more similar to the query as expected. For this query example in the BUNCA approach, one of the retrievals is wrong, but it is interpretable. The second writing style of digit "4" is very similar to digit "9" and can be morphed into it by a slight change.

\subsubsection{Retrieval of Histopathology Patches}

Query retrieval can be very useful for histopathology data in hospitals where similar patches are extracted from the database to rely on already diagnosed cases. The type of disease or tissue can be found out by a majority vote amongst the retrievals \cite{kalra2020pan}. Fig.  \ref{figure_query} shows retrievals for two different tissue types, which are tumor and mucus. The former has more complex patterns, in contrast to the latter one. As the figure shows, the retrievals are very similar to the pattern of query patch. 

\setlength{\tabcolsep}{5pt}
\begin{table}[!t]
\caption{Comparison of the proposed triplet mining approaches with the baselines on the MNIST dataset.}
\label{table_mnist}
\centering
\scalebox{1}{
\begin{tabular}{l|cccc}
\hline\hline
                                                   &  R@1                       & R@4                       & R@8                       & R@16                       \\ \hline
BA \cite{ding2015deep} & 79.31 & 93.53 & 96.55 & 98.21 \\ \hline
BSH \cite{schroff2015facenet}  & 78.95 & 92.61 & 96.09 & 98.17 \\ \hline
BH \cite{hermans2017defense} & 85.75 & 95.31 & 97.43 & 98.63              \\ \hline
EP \cite{xuan2020improved} & 73.34 & 90.09 & 95.08 & 97.68              \\ \hline
DWS \cite{wu2017sampling} & 76.44 & 91.35 & 95.72 & 97.68               \\ \hline
NCA \cite{goldberger2005neighbourhood} & 85.40 & 95.48 & 97.46 & 98.76      \\ \hline
proxy-NCA \cite{movshovitz2017no} & 83.71 & 94.69 & 97.31 & 98.55  \\ \hline\hline
BUT & 88.03 & 96.25 & 98.15 & 99.09  \\ \hline
BUNCA & 78.67 & 92.44 & 95.77 & 98.02  \\ \hline\hline
\end{tabular}
}
\end{table}

\subsection{Comparison with Baseline Methods}

In Tables \ref{table_mnist} and \ref{table_CRC}, we compare the proposed BUT and BUNCA approaches with the existing triplet mining methods in the literature.
These tables report the Recall@$k$ (R@$k$) metric on the embedded test data, for different values of  $k$.
The baseline approaches, which we compare with, are BA \cite{ding2015deep}, BSH \cite{schroff2015facenet}, BH \cite{hermans2017defense}, EP \cite{xuan2020improved}, DWS \cite{wu2017sampling}, NCA \cite{goldberger2005neighbourhood}, and proxy-NCA \cite{movshovitz2017no}; these methods were briefly introduced in Section \ref{section_introduction}.
Among these methods, DWS is a sampling method that samples from the existing instances in the mini-batch in contrast to our proposed approach, which samples from the distribution of data. 

Table \ref{table_mnist} reports the results for the MNIST dataset. The proposed BUT approach outperforms all other methods. Moreover, BUNCA performs better than EP and DWS, where DWS is also a sampling approach for triplet mining.
The results for the CRC histopathology data are reported in Table \ref{table_CRC}. On this data, the performance of BUNCA is closer to BUT. In most cases, BUT has the best performance against all the baseline approaches. On this dataset, BUNCA performs better than BA, BSH, EP, DWS, NCA, and is comparable with proxy-NCA. Overall, these two tables demonstrate the effectiveness of the proposed mining approaches for triplet training.

\setlength{\tabcolsep}{5pt}
\begin{table}[!t]
\caption{Comparison of the proposed triplet mining approaches with the baselines on the CRC dataset.}
\label{table_CRC}
\centering
\scalebox{1}{
\begin{tabular}{l|cccc}
\hline\hline
                                                   &  R@1                       & R@4                       & R@8                       & R@16                     \\ \hline
BA \cite{ding2015deep}                                                                                     & 38.54 & 66.76 & 80.64 & 89.97\\ \hline
BSH \cite{schroff2015facenet}                                   & 30.85                     & 60.39                     & 77.73                     & 90.33          \\ \hline
BH \cite{hermans2017defense}                     & 79.09                     & 92.60                     & 96.00                     & 97.95    \\ \hline
EP \cite{xuan2020improved}                                                                            & 69.94                     & 87.88                     & 93.20                     & 96.38  \\ \hline
DWS \cite{wu2017sampling}                      & 76.06                     & 91.31                     & 95.34                     & 97.58   \\ \hline
NCA \cite{goldberger2005neighbourhood}                                                                        &   77.87   &   92.25   &   95.92    &  98.01 \\ \hline
proxy-NCA \cite{movshovitz2017no}                                                               & 78.85 & 92.24 & 95.80 & 97.78 \\ \hline\hline
BUT                                                                      &  79.14 & 92.32 & 95.60 & 97.65 \\ \hline
BUNCA                                                                    & 78.67  & 92.28 & 95.64 & 97.71 \\ \hline\hline
\end{tabular}
}
\end{table}

\section{Conclusions and Future Direction}\label{section_conclusion}

Different triplet mining approaches have been proposed since the introduction of triplet networks. In this paper, we proposed a triplet mining method which considers a multivariate normal distribution for the embedding of every class through sampling the triplets from these distributions rather than from the existing instances in the mini-batch. By Bayesian updating, the distributions are dynamically updated using the received stream of mini-batches. This approach makes use of the stochastic information of the embedding space, rather than being restricted to the existing instances, for better discrimination of classes. The proposed BUT and BUNCA approaches of the dynamic triplet sampling were validated by experiments on two public  datasets and compared against baseline methods from literature. As a possible future work, one can explore a mixture of Gaussian distributions for every class of data using expectation maximization.

\bibliographystyle{IEEEtran}

\bibliography{references}

\end{document}